# Laser Data Based Automatic Generation of Lane-Level Road Map for Intelligent Vehicles

Zehai Yu, Hui Zhu, Linglong Lin, Huawei Liang, Biao Yu, Weixin Huang

*Abstract*—With the development of intelligent vehicle systems, a high-precision road map is increasingly needed in many aspects. The automatic lane lines extraction and modeling are the most essential steps for the generation of a precise lane-level road map. In this paper, an automatic lane-level road map generation system is proposed. To extract the road markings on the ground, the multi-region Otsu thresholding method is applied, which calculates the intensity value of laser data that maximizes the variance between background and road markings. The extracted road marking points are then projected to the raster image and clustered using a two-stage clustering algorithm. Lane lines are subsequently recognized from these clusters by the shape features of their minimum bounding rectangle. To ensure the storage efficiency of the map, the lane lines are approximated to cubic polynomial curves using a Bayesian estimation approach. The proposed lane-level road map generation system has been tested on urban and expressway conditions in Hefei, China. The experimental results on the datasets show that our method can achieve excellent extraction and clustering effect, and the fitted lines can reach a high position accuracy with an error of less than 10 cm.

## I. INTRODUCTION

A high definition road map is significant for high-level intelligent vehicle systems. As the most essential elements of the road map, lines on the traffic lane play an important role in intelligent vehicle's different tasks, e.g., lane keeping, lane-level path planning, and high-precision ego-vehicle localization. In the process of traditional digital map construction for intelligent vehicles, the methods based on the satellite image and aerial image are applied [1][2]. These methods obtain road collection information by manual labeling or basic image processing. However, due to the insufficient resolution of satellite images or aerial images, it is difficult to provide lane level road information. And it is increasingly unable to meet the needs of intelligent vehicles.

In order to obtain road maps with higher accuracy and richer information, the method of combining perception sensors with high-precision vehicle positioning modules has been widely used in recent years. In these methods, road markings (e.g., lane line, road arrow, and zebra crossing) are obtained through perception sensors and then combined with positioning information to construct a road map. The commonly used perception sensors are mainly cameras [3] and multi-layer LiDAR [4]-[9]. Although the vision-based solution is low in cost, it can't obtain accurate road marking position information and the overall accuracy is relatively low. And the vision-based method is easy to be affected by environmental factors such as light. In contrast, the method based on LiDAR can Overcome these problems.

Currently, there are a large number of road marking extraction and road map construction methods based on LiDAR. A road extraction method based on dense point clouds generated by vehicle-mounted laser scanning is proposed in [4] and [5]. Since the road marking has a reflection intensity that is significantly different from that of the road surface, this method realizes the segmentation of the road surface and the road marking by calculating the reflection intensity threshold. However, they only extracted road markings and didn't perform clustering and recognition processing on them.

A particle-filter-based method that uses mobile LiDAR data was proposed in [6] to extract lane lines and cluster them. However, this method only extracts the centerline of the lane, and cannot obtain lane-level road information. At the same time, the actual shape and line type information of the lane line is ignored. Therefore, such methods cannot meet the needs of high-level autonomous driving for road information.

Reference [7] realizes the extraction and clustering of lane markings points by dividing the road surface into several areas related to the vehicle trajectory. However, this method has strict requirements on the driving trajectory of the vehicle in order to get rid of the interference of road markings such as steering arrows on the lane line. And it is necessary to set parameters such as road width and lane width, so it is not very easy to be realized.

An adaptive multi-threshold method was applied in [8] to extract the road markings points and then they are clustered by Euclidean clustering method. Similarly, 错误!未找到引用源。 projected the extracted road markings to a two-dimensional raster image and clustered them by the method of region growth algorithm. Then the shape characteristics of the minimum bounding rectangle (MBR) of each cluster are calculated to identify these clusters. These methods proposed in [8]错误!未找到引用源。 show good results in common scenes. However, when it comes to the road surface with complex markings or worn markings, the clustering algorithm doesn't perform well. In order to solve the problem of lane line wear, [9]proposed a line association method to connect different small segments of the same line. However, it requires the establishment of a complex lane line model and relies on a large number of manually set parameters.

Besides, current researches basically focus on the extraction and recognition of lane elements. Few researchers discuss how to separate lane lines belonging to different lanes and construct them into parametric lane curve objects.

This paper is organized as follows. Section II introduces the road map generation system and describes the main method developed in this work. Section III details the experimental method and the results. Section IV concludes this paper and presents the future works.

## II. METHOD

### A. Overall of the System

The automatic lane-level road map generation system is mainly composed of four parts: (1) data preprocessing, (2) reflectivity based road marking extraction as shown in Fig. 1.

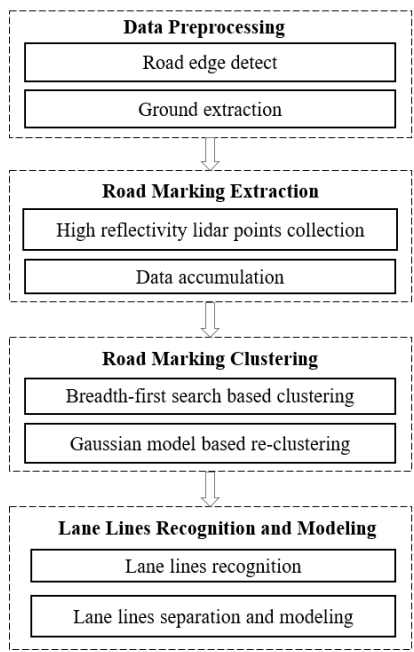

Fig. 1  Overall system architecture.

The process of data preprocessing is mainly for road boundary detection and obstacle removal. Since the algorithm proposed in this paper is mainly used for lane line extraction and modeling, curb detection is needed to determine the region of interest of the algorithm. At the same time, in order to prevent the interference of obstacles (such as vehicles and pedestrians, etc.) on the road, a suitable road extraction algorithm is also an important prerequisite.

We adopted a multi-window dynamic thresholding method to achieve stable road marking points extraction. Then, according to the position information provided by the vehicle positioning system, the point cloud data of each frame is accumulated, as shown in Fig. 2 (a).

After the extraction and accumulation of the road marking point cloud, it is necessary to apply the road marking clustering algorithm to prepare for the subsequent lane line extraction and recognition steps. In order to simplify the algorithm, we project the point cloud to a two-dimensional raster image, as shown in Fig. 2 (b). The result of the clustering step is shown in   Fig. 2 (c).

The main types of lane lines are dashed lines and solid lines. Dashed lines can be easily identified by the shape parameters of their minimum bounding rectangle (MBR). For solid lines, we calculate their eigenvector in the sliding window to distinguish. The result of line recognition is shown in Fig. 2(d).

After the above steps, lane lines of different types can be obtained. However, these lines represented by a sequence of points is inefficient in storage and it is not able to directly obtain some important line information such as the curvature and direction of the line. For this reason, we propose to use cubic spline curves to represent these lane lines and use the Bayesian estimation method to obtain more accurate curve parameters. The curve fitting result is shown in Fig. 2 (e).

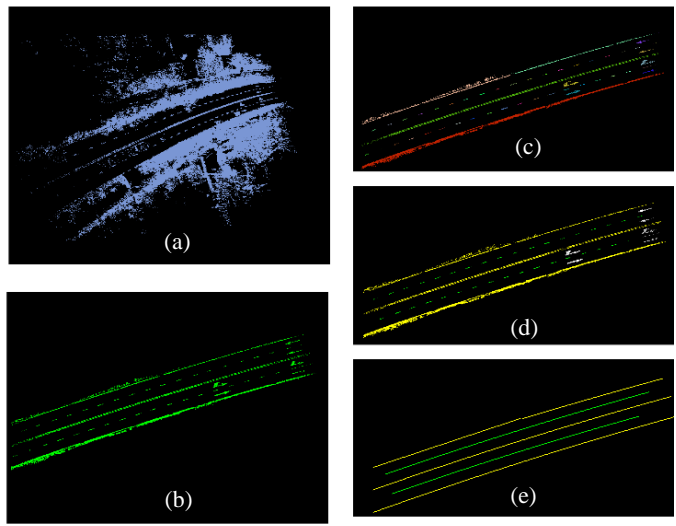

Fig. 2  Process of lane-level road map generation. (a) Road marking extraction and point cloud accumulation. (b) Point cloud rasterization (non-road points have been removed) (c) Road marking clustering (d) Lane lines recognition (e) Curve fitting.

### B. Data Preprocessing

Our algorithm focus on the road markings on the ground between the road boundaries.

We set the nearest and farthest scan line and detect the road boundaries to limit the detection area, as shown in Fig. 3.

We refer to [11] for road boundary extraction. The principle is that obstacles can be detected by analyzing the distance between consecutive laser scan rings. We first analyze the distance between adjacent ring points. If the distance is lower than the threshold we set, those points are then classified as road boundary candidates.

However, this method can cause a large number of misclassification. To remove those false positives, we apply three filters mentioned in [12]: differential filter, distance filter and regression filter.

In addition, obstacles such as vehicles, pedestrians, etc. may appear in the above detection area, making it difficult to extract road markings correctly. Therefore, we need to apply a ground segmentation algorithm to filter out non-ground points. There have been a lot of literature studies on ground segmentation. We refer to the algorithm proposed in [13], which divides the scan data into fixed-size grids, and distinguishes obstacles from the ground according to the distance difference between two adjacent rings. And the average height and variance of each grid are calculated for optimization.

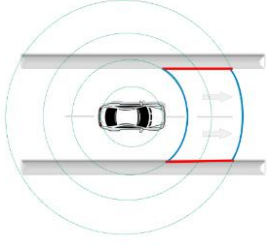

Fig. 3  Detection area.

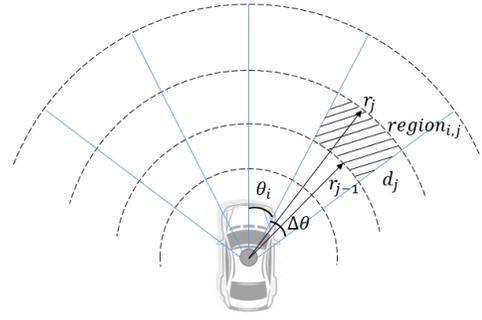

Fig. 4  Subarea division.

*C. Reflection Intensity Based Road Marking Extraction*

Since the laser points projected on the pavement material have a reflection intensity significantly different from that of the road markings, an intensity threshold can be set to realize the extraction of the point cloud of the pavement marking. A scan point with a larger reflection intensity threshold is considered to be a road marking point.

Different from the conventional reflection intensity correction method, we apply a dynamic threshold method based on multiple regions to extract the road markings. The specific process is as follows:

First, the detection area is divided into multiple fan-shaped sub-areas, as shown in Fig. 4. This fan-shaped region division mechanism can ensure that there are enough point clouds in the distant regions.

Subsequently, For each sub-area, a fixed rough intensity threshold was set to extract the road marking points. This step will cause a lot of false points. So the maximum between-cluster variance method was applied to extract the road marking points precisely.

The between-cluster variance is defined as (1).
$$\sigma_B = P_1(m_1 - m_G)^2 + P_2(m_2 - m_G)^2 \quad (1)$$
where, $p_1$ and $p_2$ denote the ratio of the road marking points and interference points to the points in the area respectively. $m_1$ and $m_2$ denote the average reflection intensity of road marking points and interference points respectively. $m_G$ denote the reflection intensity threshold.

Therefore, the reflection intensity threshold can be calculated by maximizing the between-class variance as (2).
$$\arg\max_{m_g} \sigma_B(m_g) \quad (2)$$

To extract complete road markings and classify them, we need to accumulate single-frame point cloud data based on vehicle position information to ensure sufficient point density, as shown in Fig. 2 (a). In order to facilitate the application of subsequent algorithms, we need to project the dense 3D laser point cloud to the 2D raster image, as shown in Fig. 2 (b).

*D. Road marking clustering*

We used a two-stage clustering algorithm to cluster the lane markings on the ground.

First, the pre-clustering algorithm based on breadth-first search was applied on the extracted road marking points. For each point in the raster image, the search region was set to k ∗ k size. It should be noted that there is often a short distance between the road markings on the actual road, such as the steering arrow and the lane line. Therefore, in order to prevent them from being divided into the same type of clusters, it is necessary to set a smaller search area size. The pre-clustering result is shown in Fig. 5.

However, for some worn road markings, because of incomplete extraction of marking points, this step will cause an over-segmentation problem, that is, the same mark is misclassified into multiple marks, as shown in the red circle in Fig. 5 (a).

To solve this problem, a re-clustering algorithm based on Gaussian kernel transformation is used to merge the over-segmented markings. The idea of the algorithm is to construct an objective function describing the degree of deviation of a certain point of class A relative to another certain point of class B. And Gaussian kernel transformation is performed to satisfy the Gaussian distribution model, and then the possibility that the point of class A belongs to class B is calculated.

Suppose that all points of a certain cluster form a point set M. For the point $M_i$ in this point set, create a circle with it as the center. Suppose the set of points in the circle belonging to this cluster is $m = \{m_1, m_2, ..., m_k\}$, and the set of points not belonging to this cluster is $n = \{n_1, n_2, ..., n_l\}$.

We describe the deviation of $n_i$ and $m_j$ by (3).
$$p = \left|\frac{t_{ni} - t_{mj}}{l_{ni} - l_{mj}}\right| \quad (3)$$
where, $t_{ni} - t_{mj}$ denotes the distance between these two points in the transverse direction of the road and $l_{ni} - l_{mj}$ denotes the distance in the driving direction.

Perform Gaussian kernel transformation on $p$ to make the mapped value range meet the requirements of probability density function, and construct the objective function as (4).
$$p(n_i|m_j) = exp(-p^2/2\sigma^2)/\sqrt{2\pi}\sigma \quad (4)$$

where, $\sigma = \sqrt{\frac{\sum_{i=1}^{w}|(t_{qi} - t_{mj})/(l_{qi} - l_{mj})|}{w-1}}$ and $Q = \{q_1, q_2, ..., q_w\}$

denotes all the points in the circle of $m_j$.

Equation (5) can be obtained from the conditional probability formula.

where $p(m_j) = 1/k$.

The point $n_i$ will be set to cluster $m$ if $p(n_i) > k \cdot \frac{1}{\sqrt{2\pi}\sigma}$ and then cluster $m$ and cluster n will be merged. The re-clustering result is shown in Fig. 5 (b).

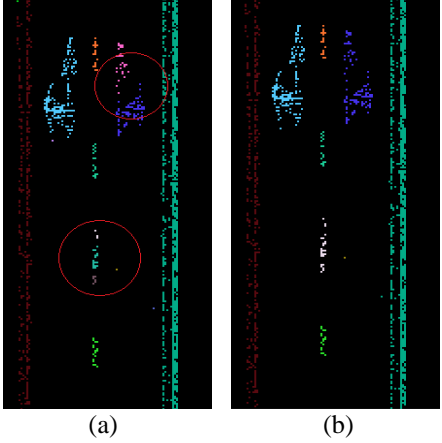

(a)        (b)

Fig. 5 Two-stage clustering.

*E. Lane lines recognition and modeling*

*1) Lane lines recognition*

After the extraction and clustering of road markings, another important step is to identify these clusters. For each cluster, we calculate the shape characteristics of the MBR, such as length, width, and aspect ratio.

The dash line has a smaller width and a larger aspect ratio, so it can be easily detected.

The MBR of a solid line generally has a large scale in at least one direction and a large aspect ratio, so it is also easy to detect. However, there is a problem with solid line detection: in scenes such as intersections, there are often situations where the stop line is connected to the solid line, as shown in Fig. 6 (a).

In order to distinguish between the solid line and the stop line, we first calculate the eigenvector of the dashed line. For each dash line, extract the midpoint and calculate the covariance matrix of the midpoint relative to other points as (5).

$$C_{P_m} = \frac{1}{n}\sum_{k=1}^{n}(p_i - p_m) \cdot (p_i - p_m)^T \quad (5)$$

Through the eigenvalue decomposition on the covariance matrix, the distribution feature of $p_m$ is set as the eigenvector corresponding to the maximum eigenvalue, and is denoted as $v_{p_m}$.

In order to distinguish between the solid line and the stop line, we sample a sequence of points and create a window for each point, as shown in Fig. 6 (b). Then the eigenvector for each sampled point in its window is calculated and denoted as $v_{p_s}$.

For each sampled point $p_s$, if the angle between the direction of its eigenvector $v_{p_s}$ and the eigenvector $v_{p_m}$ of the adjacent dashed line satisfies (6), then the points in the window are set as solid line points, as shown in Fig. 6 (b) window B. If this angle satisfies the following (7), the points in the window are set as stop line points, as shown in Fig. 6 (b) window A.

$$\alpha_d < |\cos\langle v_{p_s}, v_{p_m}\rangle| < \alpha_u \quad (6)$$

$$\beta_d < |\cos\langle v_{p_s}, v_{p_m}\rangle| < \beta_u \quad (7)$$

where, $\alpha_d, \alpha_u, \beta_d, \beta_u$ are manually set thresholds.

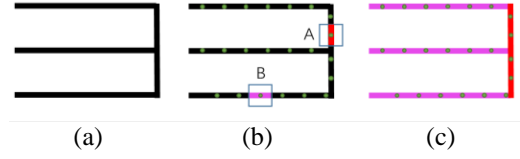

(a)        (b)        (c)

Fig. 6 Classification of solid and stop lines.

*2) Lane lines separation and modeling*

After the identification of the lane lines, there is another key issue: lane lines separation, that is, how to separate the lane lines (especially the dashed line) belonging to different lanes and construct complete lane objects. To accomplish this task, we proposed a lane lines separation algorithm based on the area prediction method.

The algorithm flow is as follows:

For a specific dashed type lane line $L_{i-1}$, we have calculated its eigenvector $v_{p_m}$ in the previous step. Then, the center point $\bar{p}_i$ of the prediction area can be calculated by (8).

$$\bar{p}_i = p_{i-1} + d_0 v_{p_m}/|v_{p_m}| \quad (8)$$

where $d_0$ is the distance between the center points of two adjacent dashed lines on the same lane, which is a known value.

Assuming that the width and length of each lane line obey the Gaussian distribution, calculate the average width and average length of the N dashed lines in the current area and their variance. Then, the length and width of the prediction area can be calculated using (9) and (10).

$$w_i = \frac{1}{N}\sum_{k=1}^{N} w_k + \lambda\sigma_w + w_\theta \quad (9)$$

$$h_i = \frac{1}{N}\sum_{k=1}^{N} h_k + \lambda\sigma_h + h_\theta \quad (10)$$

where $\sigma_w, \sigma_h$ is the standard deviation of lane line width and length distribution respectively.

According to the prediction area center point $\bar{p}_i$ and its width and length information, the prediction area of the lane line $L_i$ can be obtained, as shown in the blue area in Fig. 7. We use the parameter $P_{i, i-1}$ as the evaluation parameter to evaluate whether $L_i$ and $L_{i-1}$ belong to the same lane object.

$$P_{i, i-1} = (2N_{mi}N_{bi} - N_{bi}^2)/N_{mi}^2 \quad (11)$$

The method of using a sequence of points to represent the lane line requires a large amount of storage space and is not easy to use. Therefore, we propose a road modeling method to represent each set of lane lines as mathematical curves, as shown in (12).

$$L(s) = \sum_{t=1}^{T} L_t(s) \quad (12)$$

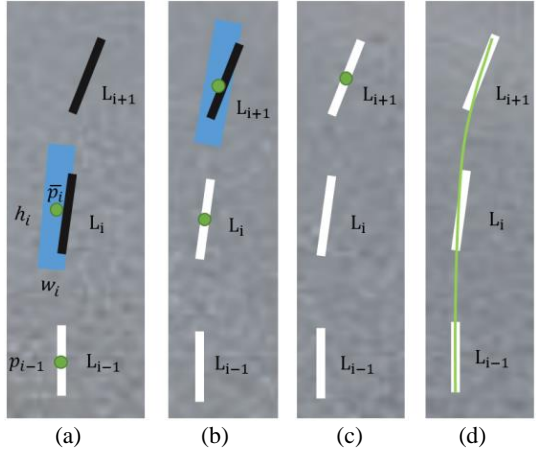

Fig. 7 Complete line object obtainment.

For each lane line, it is defined as a cubic polynomial parameter curve, as shown in (13).

$$L_t(s) = \begin{bmatrix} X_t(s) \\ Y_t(s) \end{bmatrix} = \begin{bmatrix} \sum_{r=0}^{3} P_{X_{r,m}}(s - s_t)^r \\ \sum_{r=0}^{3} P_{Y_{r,m}}(s - s_t)^r \end{bmatrix} \quad (13)$$

Therefore, the geometric data point of each lane line can be represented by only 1 parameter $s_t$ and 8 coefficients ($P_{X_{0,m}} \sim P_{X_{3,m}}, P_{Y_{0,m}} \sim P_{Y_{3,m}}$). We use the Kalman filter, a recursive Bayesian estimation method, to solve the above curve. The data points of each lane line are regarded as the observation value of the system, and the coefficients of the polynomial parameter curve are regarded as the system state we want to estimate. The method was described in detail in [7].

## III. EXPERIMENTS AND RESULTS

### A. The experimental platform and dataset

The experimental datasets were collected by the Mobile LiDAR System, which was integrated with a 128-scan-lines LiDAR, an inertial measurement unit (IMU), an odometer, and a Differential Global Position System(DGPS).

In Hefei, China, we collected two sets of data with an experimental vehicle at a speed of 30-40km/h, as shown in Fig. 8 (b) and (c), denoted as Dataset 1 and Dataset 2. Among them, data set 1 is an urban scene with a total length of about 2300m, with ordinary road surface wear, and the road markings are more complicated; data set 2 is a highway scene, with a total length of about 1700m, with simple road markings.

### B. Road Marking Extraction Result

We compared the proposed road marking extraction algorithm with the traditional fixed threshold method and the Otsu algorithm proposed in [9], as shown in Fig. 9. Our method sets different dynamic thresholds in different sub-regions, and has better extraction results in scenes with clear road markings and wear.

To evaluate the effect of the proposed algorithm, we selected five typical road scenes as shown in Fig. 10, where roads 1-3 are from dataset 1 and roads 4-5 are from dataset 2. The road marking extraction results are shown in Fig. 10 (a).

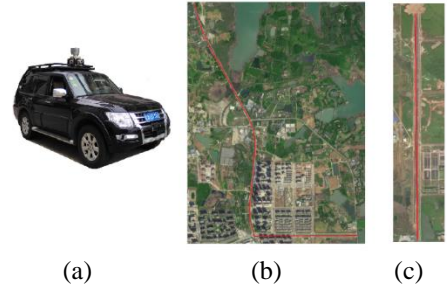

Fig. 8 Experimental equipment and sites.

### C. Lane Markings Clustering and Recognition Result

We applied our clustering and recognition algorithm in the above scenes. The results of the proposed clustering and recognition algorithm are shown in Fig. 10 (b), where the green denotes dashed lines, the magenta denotes solid lines, the red denotes stop lines and the blue denotes other elements.

| | Fixed Threshold | Otsu | Proposed | Reference |
|---|---|---|---|---|
| 1 | | | | |
| 2 | | | | |
| 3 | | | | |
| 4 | | | | |

Fig. 9 Road marking extraction.

In order to evaluate the accuracy of road marking clustering and recognition algorithm, we compared the final recognition results with the manually marked results, and define the three indicators in (14) for evaluation.

$$\begin{cases} \text{Precision} = TP/(TP + FP) \\ \text{Recall} = TP/(TP + FN) \\ \text{Fsocre} = Precision \cdot Recall/(Precision + Recall) \end{cases} \quad (14)$$

where TP, FP and FN are the number of true positives, false positives and false negatives, respectively

The statistical results are shown in TABLE I. In TABLE I, we manually counted the number of various elements in data set 1 including the dashed line, the solid line, the stop line and other elements. We compared our result with the manual inspection result and calculated the value of TP, FP and FN.

TABLE II showed the evaluation indicators related to TABLE I. The precision of datasets 1 and 2 are both over 90%. The recall rate in dataset 1 is relatively lower because some road markings on the roads are severely worn and difficult to identify.

TABLE I
STATISTICAL RESULT OF THE RECOGNITION ALGORITHM

| Data | Line type | Object | | Assessment result | | |
|---|---|---|---|---|---|---|
| | | Manual inspection | Proposed | TP | FP | FN |
| 1 | Dashed line | 559 | 512 | 485 | 27 | 74 |
| | Solid line | 128 | 132 | 120 | 12 | 8 |
| | Stop line | 14 | 14 | 13 | 1 | 0 |
| | Others | 119 | 110 | 101 | 9 | 18 |
| 2 | Dashed line | 557 | 554 | 538 | 16 | 19 |
| | Solid line | 69 | 72 | 65 | 7 | 4 |
| | Stop line | 1 | 1 | 1 | 0 | 0 |
| | Others | 125 | 135 | 110 | 25 | 15 |

TABLE II
EVALUATION INDICATORS OF THE RECOGNITION ALGORITHM

| Dataset | Precision | Recall | F-score |
|---|---|---|---|
| 1 | 0.936 | 0.878 | 0.453 |
| 2 | 0.937 | 0.950 | 0.472 |

*D. Line Fitting Result*

We applied the proposed line separation and modeling method, then the detected lines are fitted using the calculated parameters, as shown in Fig. 10 (c). The letter B and S denote the dashed line and the solid line, respectively. Obviously these lane lines were accurately separated and numbered, which is essential for the usability of the road map.

Because the extracted lines have abnormal points or error clustering, the fitted curves have errors compared with their actual positions. In order to evaluate the accuracy of the fitted curve, we manually marked some dashed and solid lines, and defined the Root Mean Square Error (RMSE) as the evaluation index.

We manually label the lines of some areas in dataset 1 and calculate the error of the fitted curve. The fitting result of the dashed lines is shown in Fig. 11, with an average error at 5.4607 cm. The fitting result of the solid lines is shown in Fig. 12, with an average error at 8.3719 cm.

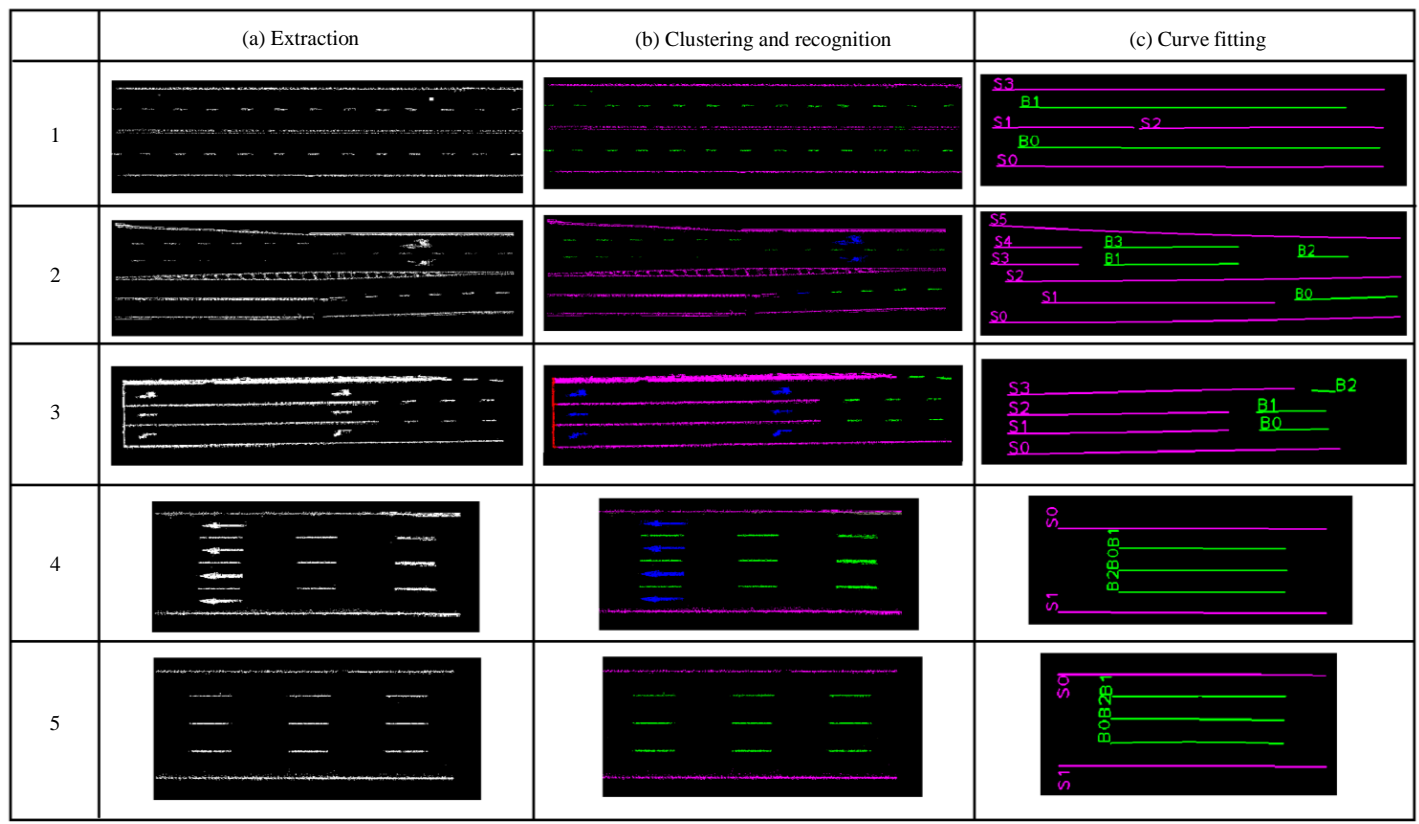

Fig. 10 Clustering and recognition result.

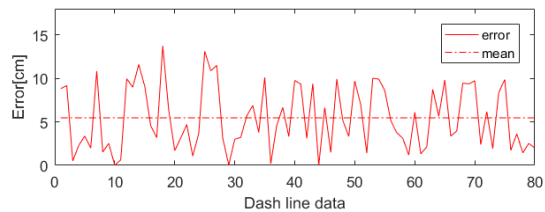

Fig. 11 Fitting error of the dashed lines.

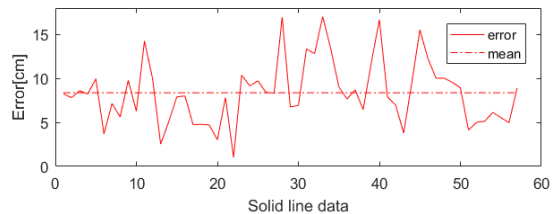

Fig. 12 Fitting error of the solid lines.

IV. CONCLUSION

In this paper, an automatic lane markings extraction, recognition and fitting algorithm is implemented to generate a high-precision road map. The experimental results show promising performance. However, our method does not perform well in places where road markings are extremely worn because the high reflectivity coating is almost completely lost. How to use a suitable predictive model to obtain the lane lines in more challenging scenarios will be our main task in the next works.

ACKNOWLEDGMENT

This work was supported by National Key Research and Development Program of China (Nos. 2016YFD0701401, 2017YFD0700303 and 2018YFD0700602), Equipment Pre-research Program (Grant No. 301060603), Youth Innovation Promotion Association of the Chinese Academy of Sciences (Grant No. 2017488), Key Supported Project in the Thirteenth Five-year Plan of Hefei Institutes of Physical Science, Chinese Academy of Sciences(Grant No.KP-2017-35, KP-2017-13,KP-2019-16), Independent Research Project of Research Institute of Robotics and Intelligent Manufacturing Innovation, Chinese Academy of Sciences(Grant No. C2018005), and Technological Innovation Project for New Energy and Intelligent Networked Automobile Industry of Anhui Province.